\documentclass{article}
\usepackage{spconf,amsmath,graphicx}
\usepackage{algpseudocode}
\usepackage{algorithm}
\usepackage{mathtools, nccmath}

\title{LANGUAGE-UNIVERSAL ADAPTER LEARNING WITH KNOWLEDGE DISTILLATION FOR END-TO-END MULTILINGUAL SPEECH RECOGNITION}
\name{Zhijie Shen, Wu Guo, Bin Gu\thanks{Thanks to the National Natural Science Foundation of China (Grant No. U1836219) for funding.}}
\address{The Department of Electronic Engineering and Information Science (EEIS)\\
University of Science and Technology of China\\
Hefei, China
}
\begin{document}
%
\maketitle
\begin{abstract} In this paper, we propose a language-universal adapter learning framework based on a pre-trained model for end-to-end multilingual automatic speech recognition (ASR). For acoustic modeling,  the wav2vec 2.0 pre-trained model is fine-tuned by inserting language-specific and language-universal adapters. An online knowledge distillation is then used to enable the language-universal adapters to learn both language-specific and universal features. The linguistic information confusion is also reduced by leveraging language identifiers (LIDs). With LIDs we perform a position-wise modification on the multi-head attention outputs. In the inference procedure, the language-specific adapters are removed while the language-universal adapters are kept activated. The proposed method improves the recognition accuracy and addresses the linear increase of the number of adapters’ parameters with the number of languages in common multilingual ASR systems. Experiments on the BABEL dataset confirm the effectiveness of the proposed framework. Compared to the conventional multilingual model, a 3.3\% absolute error rate reduction is achieved. The code is available at: https://github.com/shen9712/UniversalAdapterLearning.
\end{abstract}
\section{Introduction}
\label{sec:intro}

With the widespread of end-to-end automatic speech recognition (ASR) frameworks \cite{dong2020cif, kahn2020self, miao2020transformer}, multilingual ASR has become a research hotspot. This is mainly due to easy training and deployment procedures in real-world applications, especially in low-resourced scenarios.

One of the key challenges in multilingual ASR is language confusion. The most intuitive idea to address this issue is using language-specific parameters in the model training \cite{yi2018language, gaur2021mixture, kannan2019large, winata2020adapt}. For instance, in \cite{yi2018language}, the lower LSTM layers were shared among multiple languages as a common feature extractor, whereas the upper layers were language specific. \cite{gaur2021mixture} introduced the informed mixture-of-experts layers in which each expert was assigned to one language.

The adapter-based modeling technique has been successfully used for domain adaptation in computer vision \cite{sung2022vl}, natural language processing \cite{pfeiffer2020unks}, and machine translation \cite{bapna2019simple}. Adapters make effectively domain-specific (language-specific in our case) adjustments to the activations in a network. In multilingual ASR, \cite{kannan2019large} investigated using adapter modules for nine Indian languages in an RNN-T model. Furthermore, The Adapter-and-Adjust framework was introduced in \cite{winata2020adapt}, where both language-specific (LSA) and common adapters were applied to an encoder-decoder network. The LSA was focused on adapting the shared network weights to a particular language, whereas the common adapter was used to learn shared knowledge. Nevertheless, the number of LSA’s parameters grows linearly with the number of languages, which limits large-scale multilingual modeling. Also, the common adapter is trained based on imbalanced multilingual data, hence it is biased towards dominant languages.

To address the above-mentioned issues, we propose to merge the language-specific and language-agnostic information into one language-universal adapter (LUA). In our approach, the LSA and LUA are first inserted into the wav2vec 2.0 pre-trained model \cite{baevski2020wav2vec} in the training procedure. The wav2vec 2.0 pre-trained model is used due to its high performance in low-resourced ASR. The LSA captures language-specific features which are then transferred to LUA through online knowledge distillation (KD). This results in the improved robustness of LUA to data imbalance and domain (language) shifts. Note that only LUA is used for inference.

It is generally acknowledged that incorporating the language identifier (LID) is beneficial for multilingual ASR. For instance, \cite{shetty2020improving, zhou2022configurable, zhu2020multilingual} showed that training and testing conditioned on LID improves the performance and reduces language confusions. LID can be also applied in different positions of the network. For example, \cite{zhou2022configurable} simply concatenated a one-hot vector to the input features of the encoder network, whereas \cite{zhu2020multilingual} concatenated  LID to multi-head attention inputs in the transformer model.

In this paper, in order to further reduce language confusion, we propose to use LID as multi-head attention prefixes. This is inspired by the prefix-tuning method \cite{li2021prefix}. This method performs a position-wise modification on the multi-head attention outputs, which is more effective than injecting LID into the input features or hidden features.

\label{sec:format}
\begin{figure}[t]
    \centering
    \includegraphics[width=0.49\textwidth]{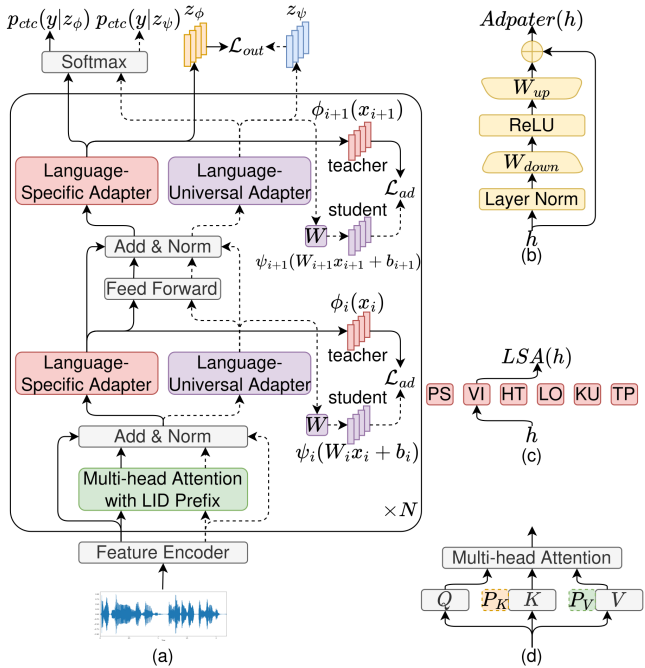}
    \caption{The language-universal adapter learning framework: (a) The adapter modules are inserted in transformer layers. The final loss contains the adapter-based distillation loss, output-based distillation loss, CTC loss (with LSA), and CTC loss (with LUA); (b) Each adapter consists of a layer normalization, down-projection, non-linearity, and up-projection; (c) LSA, where the corresponding adapter is activated given the input's language; (d) Multi-head attention with LID prefixes.}
    \label{fig:framework}
\end{figure}

\section{Methods}

Figure 1 shows the training framework of the proposed method. It is built on a wav2vec 2.0 pre-trained model with two key novel contributions: (1) Language-universal adapter, and (2) Multi-head attention with LID prefixes.

\subsection{Fine-tuning the wav2vec 2.0 model for ASR}

Wav2vec 2.0 is a transformer-based model trained to extract contextualized representations from raw audio signals \cite{baevski2020wav2vec}. This model consists of three sub-modules including a feature encoder, transformer encoder, and quantization module. Feature encoder is a multi-layer CNN that processes the input signal into low-level features. Based on this representation, the transformer module is further applied to produce contextualized representation. The quantization module discretizes the low-level features into a trainable codebook. To train the model, parts of the low-level features are masked from the transformer module, and the objective is to identify the quantized version of the masked features based on its context.

The objective of the wav2vec 2.0 is to use the learned representations to improve ASR performance. It requires less data for training, allowing its application for low-resourced ASR. To this end, the model trained as described above is fine-tuned for ASR with speech-transcription paired data. We build a contrastive system using the wav2vec 2.0 pre-trained model, where a randomly initialized linear projection layer is added on top of the contextual encoder and the Connectionist Temporal Classification (CTC) loss \cite{graves2006connectionist} is minimized.

\subsection{Language-universal adapter}

In the proposed method, adapters are inserted in each transformer layer. Similar to \cite{winata2020adapt}, we use two types of adapter modules including the Language-Specific Adapter (LSA) and Language-Universal Adapter (LUA) (Fig. 1(a)), These adapters are inserted next to the multi-head attention and feed-forward blocks. LSA contains separate parameters for each language (Fig. 1(c)), whereas LUA is a single standard adapter and it is shared across languages.

In \cite{winata2020adapt}, the summation of the adapters' outputs is considered the final adapter output. Here, however, we use online knowledge distillation to transfer knowledge from LSA to LUA. This enables LUA to learn both common features and language-specific features (see Algorithm 1). Each training batch consists of two forward passes and one backward pass. In the forward passes, LSA and LUA are activated to produce language-specific and language-agnostic features, respectively. In the backward pass, the model is optimized by minimizing the CTC loss corresponding to each forward pass. In addition to the CTC losses, the mean squared error (MSE) loss between the LSA and LUA feature maps is obtained for knowledge distillation:
%
\begin{equation}
\label{eq:ad}
    \mathcal{L}_{ad} = \frac{1}{P}\sum_{i=1}^{P}  {  \mbox{MSE}(\phi_i(\mathbf{x}_i), \psi_i(\mathbf{W}_i\mathbf{x}_i+\mathbf{b}_i))   }
\end{equation} 
where $P$ is the number of positions to insert adapters ($P$ is equal to twice the number of layers), $\phi_i$ denotes LSA, and $\psi_i$ denotes LUA. An additional linear layer ($\mathbf{W}_i$ and $\mathbf{b}_i$) is also  introduced after LUA and before loss calculation. This is because for $L$ languages LSA has $L$ times as many parameters as LUA and it is difficult for LUA to directly learn from LSA.

In addition to $\mathcal{L}_{ad}$, the MSE loss of the predicted logits of the output layer is used for further knowledge distillation:
%
\begin{equation}
\label{eq:out}
     \mathcal{L}_{out} = \mbox{MSE}(\mathbf{z}_{\phi},\mathbf{z}_{\psi})
\end{equation}
where $\mathbf{z}_{\phi}$ and $\mathbf{z}_{\psi}$ are the predicted logits of using the backbone model and LSA, and the backbone model and LUA, respectively. Finally, the combination of the CTC losses and the above-mentioned auxiliary loss is used to train the model:
\begin{equation}
\label{eq:mt}
    \mathcal{L}_{mt} \!= \!-\!\log{\!p_{ctc}(\mathbf{y}|\mathbf{z}_{\phi})} \!-\! \log{\!p_{ctc}(\mathbf{y}|\mathbf{z}_{\psi})} \!+\! \alpha \mathcal{L}_{ad} \!+\! \beta \mathcal{L}_{out}
\end{equation}
For decoding, the LSA and its corresponding linear layer are dropped and only the backbone model and LUA are used.

\begin{algorithm}[hbt!]
\caption{Learning Language Universal Adapter}\label{alg:KD}
\begin{algorithmic}
\Require{$D$: multilingual data}
\Require{$\theta$, $\phi$ and $\psi$: backbone model, LSA and LUA}
\Require{$\alpha$, $\beta$ and $\lambda$: step size hyperparameters}
 \State Randomly initialize $\theta$, $\phi$ and $\psi$;
 \State Copy wav2vec 2.0 pre-trained encoder parameters in $\theta$
 \While{not done}
   \State Sample batch of multilingual utterances $x \sim D$
   \State Compute $ \mathbf{z}_{\phi} $ and $ \phi_i(\mathbf{x}_i) $ by $\theta$ and $\phi$ forwarding
  \State Compute $\mathbf{z}_{\psi} $ and $ \psi_i(\mathbf{W}_i\mathbf{x}_i+\mathbf{b}_i) $ by $\theta$ and $\psi$ forwarding
  \State Compute CTC posteriors $p_{ctc}(\mathbf{y}|\mathbf{z}_{\phi})$ and $p_{ctc}(\mathbf{y}|\mathbf{z}_{\psi})$
  \State Compute adapter-based distillation loss $\mathcal{L}_{ad}$ in (\ref{eq:ad})
  \State Compute output-based ditillation loss $\mathcal{L}_{out}$ in (\ref{eq:out})
  \State Compute the multi-task loss $\mathcal{L}_{mt}$ using $p_{ctc}(\mathbf{y}|\mathbf{z}_{\phi})$, $p_{ctc}(\mathbf{y}|\mathbf{z}_{\psi})$,  $\mathcal{L}_{ad}$ ,  $\mathcal{L}_{out}$ , $\alpha$ and $\beta$ in (\ref{eq:mt})
  \State Update model $u \!\gets\! u - \lambda \bigtriangledown_{u}\mathcal{L}_{mt},$ where  $u\in{\theta, \phi \mbox{ and } \psi}$
 \EndWhile
\end{algorithmic}
\end{algorithm}

\subsection{Multi-head attention with LID prefixes}

Here, we propose a novel method to leverage LID by prepending the LID embedding vectors to the multi-head attention keys and values. In this approach , for each training sample, LID is firstly represented as one-hot vectors. It is then parametrized by an embedding layer followed by two linear layers with $\mbox{tanh}$ activations, producing the prefixes:
\begin{equation}
  [\mathbf{P}_k, \mathbf{P}_v] = \mathbf{W}_2 \tanh{\left( \mathbf{W}_1  \mbox{Embed}\left(LID\right) + \mathbf{b}_1\right)} + \mathbf{b}_2 
\end{equation}
Two sets of prefixes $ \mathbf{P}_k,\mathbf{P}_v \in R^d $  are concatenated with the original key and value in time. The attention is then performed on the new prefixed keys and values and the computation of each $ \mbox{head} $ becomes:
\begin{align}
 \mbox{head}  &=  \mbox{Attn}  \left( \mathbf{x}\mathbf{W}_q  ,    \mbox{concat}  \left(\mathbf{P}_k ,  \mathbf{x}\mathbf{W}_k \right)  , \mbox{concat}  \left(\mathbf{P}_v , \mathbf{x}\mathbf{W}_v  \right)   \right) \nonumber  \\[-1mm]  
 &=  \mbox{softmax}  \left(  \mathbf{x}\mathbf{W}_q{\mbox{concat}  \left(  \mathbf{P}_k,\mathbf{x}\mathbf{W}_k \right)}^T\right)  \begin{pmatrix} \mathbf{P}_v \\[-1mm]   \mathbf{x}\mathbf{W}_v \end{pmatrix}  \nonumber \\[-1mm]  
&=  \left(1-\gamma(\mathbf{x})\right)\mbox{Attn}  \left(\mathbf{x}\mathbf{W}_q,\mathbf{x}\mathbf{W}_k,\mathbf{x}\mathbf{W}_v\right) +  \nonumber \\[-1mm]
&\quad \gamma(\mathbf{x})\mbox{Attn}  \left(\mathbf{x}\mathbf{W}_q,\mathbf{P}_k,\mathbf{P}_v \right)
\end{align}
where $N_h$  denotes the number of heads, $\mathbf{P}^{(i)}_k$, $\mathbf{P}^{(i)}_v \in R^{d/N_h} $  denote the prefixes, and $ \mathbf{W}_q^{(i)}, \mathbf{W}_k^{(i)}, \mathbf{W}_v^{(i)}  $ are used to project inputs to queries, keys, and values respectively. Furthermore, $\gamma(\mathbf{x})$ is a scalar that represents the sum of normalized attention weights on the prefixes:
\begin{equation}
  \renewcommand{\arraystretch}{1}
    \! \gamma(\mathbf{x}) \! = \! \frac{ \sum_{i} \exp{\left( \mathbf{x} \mathbf{W}_q \mathbf{P}^{T}_k \right)}_i }{ \sum_{i}  \exp {   \left(  \mathbf{x} \mathbf{W}_q \mathbf{P}^{T}_k  \right)}_{i} \! + \! \sum_{j}   \exp{   \left(  \mathbf{x} \mathbf{W}_q \mathbf{W}^{T}_k  \mathbf{x}  ^T   \right)}_{j}}
\end{equation}
Therefore, this method is equivalent to performing a position-wise modification on the multi-head attention outputs. Once training is complete, the embedding and linear layers are dropped, and only $ \mathbf{P}_k$ and $ \mathbf{P}_v$ of each language are stored.

\section{EXPERIMENTAL SETUP}
\label{sec:pagestyle}
\subsection{Datasets}

We use the BABEL dataset and six languages are selected including Pashto (ps), Vietnamese (vi), Haitian (ht), Lao (lo), Kurdish (ku), and Tok Pisin (tp) \cite{gales2014speech}. The dev folder of the BABEL dataset is used as the test set since ``eval'' has not been open-sourced. Also, 10\% of the training set is used as dev data. Samples from all languages are pooled and shuffled to form the multilingual training set. All audio is resampled to 16kHz to satisfy the wav2vec 2.0 input requirements. The statistics of the dataset are presented in Table 1.

\begin{table}[t]
  \caption{Data statistics per language (in hours).}
  \label{tab:data}
  \centering
  \begin{tabular}{llll}
  \hline
     \textbf{Language}  & \textbf{Train} & \textbf{Dev} & \textbf{Test} \\
     \hline
     Pashto     &  70.6  & 7.6  & 10.0 \\
     Vietnamese &  79.1  & 8.6  & 11.0\\ 
     Haitian    &  59.8  & 7.1  & 10.6 \\
     Lao        &  58.5  & 7.0  & 10.6 \\ 
     Kurdish   &  37.4  & 4.4  & 10.2\\  
  Tok Pisin     &  35.4  & 4.0  & 10.0\\   
  \hline
  \end{tabular}
\end{table}

\subsection{Model configuration}

The backbone wav2vec 2.0 model has the same configuration as in \cite{conneau2020unsupervised}. We fine-tune the Base and Large models which were pre-trained on 960h Librispeech data and data from 53 languages, respectively. The feature encoder is frozen during the fine-tuning process. Each LSA and LUA in the same transformer layer has the same configuration. For both Base and Large models, the adapter dimension is set to 256, and adapters and LID prefixes are applied to the top 6 transformer layers. The LID prefix re-parameterization module contains two linear projections with the output dimensions of 800 and 768(Base)/1024(Large), respectively. Characters of all target languages are concatenated to create the output vocabulary for multilingual training. Extra tokens are appended to the outputs including an unknown token, a padding token, a blank token, and a mask token, yielding 285 output nodes.

For the hyper-parameters of the multi-task loss calculation, $\alpha$ and $\beta$ are set to 0.1(Base)/0.05(Large) and 0.1(Base)/0.1(Large), respectively. All models are trained with a total of 140,000 updates, a learning rate of 1e-4, and a batch size of 25,600,000 tokens using the Fairseq toolkit. For rescoring, the 4-gram KenLM \cite{heafield2011kenlm} is trained using the transcriptions from the training set for each language. Character Error Rate (CER) is given as a performance metric.

The contrastive systems are as follows. \textbf{Mono}: the pre-trained model is fine-tuned as described in Subsection 2.1, using data from each language respectively \cite{conneau2020unsupervised}. \textbf{Multi}: The model is fine-tuned directly on this training set. The training hyperparameters are the same as described in Subsection 3.2.

\section{RESULTS}

Table 2 presents the CER performance of the proposed methods and the contrastive systems. For the Base model, we first compare the monolingual (System A0) and multilingual (System A1) models. Table 2 shows that the CERs of the monolingual and multilingual systems are close to each other except for Tok Pisin. For Tok Pisin, the CER of the multilingual baseline is much worse than the monolingual baseline. This can be attributed to the imbalanced training data \cite{kannan2019large}. It is also seen that leveraging the LID prefixes in System A2 significantly reduces the CER (from 26.8\% to 24.7\% on average). Comparing  the performance of models with LUA (System A3) and LSA (System A4) shows that LSA leads to lower CER. Table 3 also indicates that LSA increases the number of parameters by 31\% (in contrast, LUA increases it by 2.5\%).

System A5 (without KD) uses LUA and LSA in training and only LUA for decoding. Our results show that System A5 is overperformed by System A4 (which uses LSA for training and decoding). However, after incorporating the KD losses $L_{ad}$ and $L_{out}$, System A7 outperforms System A4 and it achieves an absolute CER reduction of 3.3\% compared to the multilingual baseline (System A1).

Furthermore, for the Large model, the proposed framework (System B2) outperforms the multilingual baseline (System B1), with a 2.8\% absolute performance gain. The monolingual system (System B0) can obtain a slightly better result than the proposed method at a cost of a much larger model size, as shown in Table 3.

\label{sec:typestyle}

\begin{table}[ht]
  \caption{CER of the proposed framework and the baselines. Here, System A3, A5, A6 and A7 use LUA for decoding, and System A4 uses LSA for decoding.}
  \label{tab:res-cer}
  \centering
  \setlength\tabcolsep{0.3pt}
  \begin{tabular}{p{0.53cm}p{2.65cm}p{0.75cm}p{0.75cm}p{0.75cm}p{0.75cm}p{0.75cm}p{0.75cm}p{0.75cm}}
  \hline
     \multicolumn{2}{l}{\textbf{Model}}  & \textbf{ps} & \textbf{vi} & \textbf{ht} & \textbf{lo} & \textbf{ku} & \textbf{tp}  & \textbf{avg} \\ \hline
     \multicolumn{9}{l}{Base}  \\ \hline
     A0 & Mono  & 23.0 & 22.1 & 22.1 & 24.8 & 40.2 & 17.8  & 25.0 \\  \hline
     A1 & Multi & 23.6 & 22.0 & 23.5 & 23.8 & 38.5 & 29.4  & 26.8 \\ 
     A2 & A1 + LID Prefixes & 23.5 & 21.6 & 23.5 & 23.7 & 38.9 & 17.2 &  24.7  \\
     A3 & A2 + LUA   & 23.5  & 21.9  & 23.9  & 23.2 & 38.6 & 17.3 & 24.7 \\
     A4 & A2 + LSA   & 22.4   & 21.3  & 22.5  & 23.0   & 38.1  & 16.9   & 24.0         \\ 
     A5 & \begin{tabular}[c]{@{}l@{}}A2 + LSA + LUA\\ \end{tabular}  & 23.3   & 21.6  & 23.7  & 23.2   &  38.8  &  16.9  & 24.6 \\ \hline
     A6 & \begin{tabular}[c]{@{}l@{}}A5 + $\mathcal{L}_{ad}$\\ \end{tabular} & 22.6 & 21.0 & 22.8 & 23.0 & 38.7 & 16.8 & 24.2\\
     A7 &  \begin{tabular}[c]{@{}l@{}}A6 + $\mathcal{L}_{out}$\\ \end{tabular}   & \textbf{22.1} & \textbf{20.4} & \textbf{22.1} & \textbf{22.4} & \textbf{37.3} & \textbf{16.4 } & \textbf{23.5}
\\ \hline \hline
     \multicolumn{9}{l}{Large} 
\\ \hline
     B0 & Mono  & 22.1 & 20.8 & 20.5 & 22.5 & 38.6 & 15.9 & 23.4
     \\ \hline
     B1 & Multi & 23.3 & 21.0  & 22.2 & 23.5 & 38.5 & 31.0  & 26.6
     \\ 
     B2 & Proposed    & \textbf{22.0} & \textbf{20.2} & \textbf{20.9} & \textbf{22.4} & \textbf{37.1} & \textbf{20.1}  & \textbf{23.8}
\\    \hline
  \end{tabular}
\end{table}

\begin{table}[ht]
  \caption{The number of parameters of the baseline models and the proposed framework.}
  \label{tab:res-para}
  \centering
  \begin{tabular}{lllll}
  \hline
     \textbf{Model}  & \textbf{Mono} & \textbf{Multi} & \textbf{LSA} & \textbf{LUA}   \\ \hline
     Base   &  94M*6    &  94M   & 123M & 97M  \\
     Large  &  316M*6   &  316M  & 354M & 322M \\\hline
  \end{tabular}
\end{table}

\subsection{Knowledge Distillation vs. Summation}

\cite{winata2020adapt} also used LSA and LUA for multilingual ASR modeling. They take the summation from the two types of adapters as the final adapter output. In contrast, our framework is based on knowledge distillation. Table 4 compares the two methods using the Base model with an equal number of parameters. Our methods obtain an absolute CER reduction of 1.8\%.

\subsection{LID experiment}

To demonstrate the effectiveness of using the LID prefixes in the multi-head attention module, we build three contrastive systems with different LID insertion positions including Input, Attention, and Top-6.
\textbf{Input}: LID is added to the conventional acoustic feature (in our case, the transformer's first layer’s input) \cite{kannan2019large, shetty2020improving, zhou2022configurable}. 
\textbf{Top-6}: LID is added to the top-6 transformer layer inputs. 
\textbf{Attention}: LID is concatenated to the attention input feature and an additional down-projection is also introduced after concatenation \cite{zhu2020multilingual}.
%
%
The results shown in Table 5 confirm that our method results in the lowest CER.

\begin{table}[ht]
  \caption{CER of knowledge distillation-based (KD) and summation-based (Sum) adapters.}
  \label{tab:res-kd}
  \centering
  \begin{tabular}{llllllll}
  \hline
     \textbf{Model}  & \textbf{ps} & \textbf{vi} & \textbf{ht} & \textbf{lo} & \textbf{ku} & \textbf{tp}  & \textbf{avg}
     \\\hline
     KD   & \textbf{22.1} & \textbf{20.4} & \textbf{22.1} & \textbf{22.4} & \textbf{37.3} & \textbf{16.4}  &  \textbf{23.5}
     \\
     Sum   & 23.4 & 22.7 & 23.5 & 24.1 & 39.9 & 17.7  &  25.2
\\\hline
  \end{tabular}
\end{table}

\begin{table}[ht]
  \caption{CER of different approaches to leverage LID.}
  \label{tab:res-lid}
  \setlength\tabcolsep{5.3pt}
  \centering
  \begin{tabular}{llllllll}
  \hline
     \textbf{Model}  & \textbf{ps} & \textbf{vi} & \textbf{ht} & \textbf{lo} & \textbf{ku} & \textbf{tp}  & \textbf{avg}
     \\ \hline
     Prefixes   & \textbf{22.1} & \textbf{20.4} & \textbf{22.1} & \textbf{22.4} & \textbf{37.3} & \textbf{16.4}  & \textbf{23.5}
     \\
     Input    & 22.5 & 20.9 & 22.7 & 22.8 & 38.4 & 17.6  & 24.2
     \\
     Top-6  & 22.3 & 20.9 & 22.5 & 22.6 & 37.9 & 16.6 & 23.8
     \\
     Attention  & 23.3 & 21.1 & 23.4 & 23.4 & 38.7 & 17.5  & 24.6
\\ \hline
  \end{tabular}
\end{table}

\section{CONCLUSIONS}
\label{sec:conc}
This paper introduced a language-universal adapter learning framework for multilingual speech recognition, based on the online knowledge distillation and multi-head attention with LID prefixes. Experiments confirmed the proposed framework outperforms the conventional multilingual approaches.

\newpage

\bibliographystyle{IEEEbib}
\bibliography{strings,refs}

\begin{thebibliography}{10}

\bibitem{dong2020cif}
Linhao Dong and Bo~Xu,
\newblock ``Cif: Continuous integrate-and-fire for end-to-end speech
  recognition,''
\newblock in {\em 2020 IEEE International Conference on Acoustics, Speech and
  Signal Processing}. IEEE, 2020, pp. 6079--6083.

\bibitem{kahn2020self}
Jacob Kahn, Ann Lee, and Awni Hannun,
\newblock ``Self-training for end-to-end speech recognition,''
\newblock in {\em 2020 IEEE International Conference on Acoustics, Speech and
  Signal Processing}. IEEE, 2020, pp. 7084--7088.

\bibitem{miao2020transformer}
Haoran Miao, Gaofeng Cheng, Changfeng Gao, Pengyuan Zhang, and Yonghong Yan,
\newblock ``Transformer-based online ctc/attention end-to-end speech
  recognition architecture,''
\newblock in {\em 2020 IEEE International Conference on Acoustics, Speech and
  Signal Processing}. IEEE, 2020, pp. 6084--6088.

\bibitem{yi2018language}
Jiangyan Yi, Jianhua Tao, Zhengqi Wen, and Ye~Bai,
\newblock ``Language-adversarial transfer learning for low-resource speech
  recognition,''
\newblock {\em IEEE/ACM Transactions on Audio, Speech, and Language
  Processing}, vol. 27, no. 3, pp. 621--630, 2018.

\bibitem{gaur2021mixture}
Neeraj Gaur, Brian Farris, Parisa Haghani, Isabel Leal, Pedro~J Moreno, Manasa
  Prasad, Bhuvana Ramabhadran, and Yun Zhu,
\newblock ``Mixture of informed experts for multilingual speech recognition,''
\newblock in {\em 2021 IEEE International Conference on Acoustics, Speech and
  Signal Processing}. IEEE, 2021, pp. 6234--6238.

\bibitem{kannan2019large}
Anjuli Kannan, Arindrima Datta, Tara~N Sainath, Eugene Weinstein, Bhuvana
  Ramabhadran, Yonghui Wu, Ankur Bapna, Zhifeng Chen, and Seungji Lee,
\newblock ``Large-scale multilingual speech recognition with a streaming
  end-to-end model,''
\newblock {\em arXiv preprint arXiv:1909.05330}, 2019.

\bibitem{winata2020adapt}
Genta~Indra Winata, Guangsen Wang, Caiming Xiong, and Steven Hoi,
\newblock ``Adapt-and-adjust: Overcoming the long-tail problem of multilingual
  speech recognition,''
\newblock {\em arXiv preprint arXiv:2012.01687}, 2020.

\bibitem{sung2022vl}
Yi-Lin Sung, Jaemin Cho, and Mohit Bansal,
\newblock ``Vl-adapter: Parameter-efficient transfer learning for
  vision-and-language tasks,''
\newblock in {\em Proceedings of the IEEE/CVF Conference on Computer Vision and
  Pattern Recognition}, 2022, pp. 5227--5237.

\bibitem{pfeiffer2020unks}
Jonas Pfeiffer, Ivan Vuli{\'c}, Iryna Gurevych, and Sebastian Ruder,
\newblock ``Unks everywhere: Adapting multilingual language models to new
  scripts,''
\newblock {\em arXiv preprint arXiv:2012.15562}, 2020.

\bibitem{bapna2019simple}
Ankur Bapna, Naveen Arivazhagan, and Orhan Firat,
\newblock ``Simple, scalable adaptation for neural machine translation,''
\newblock {\em arXiv preprint arXiv:1909.08478}, 2019.

\bibitem{baevski2020wav2vec}
Alexei Baevski, Yuhao Zhou, Abdelrahman Mohamed, and Michael Auli,
\newblock ``wav2vec 2.0: A framework for self-supervised learning of speech
  representations,''
\newblock {\em Advances in Neural Information Processing Systems}, vol. 33, pp.
  12449--12460, 2020.

\bibitem{shetty2020improving}
Vishwas~M Shetty and Metilda Sagaya~Mary NJ,
\newblock ``Improving the performance of transformer based low resource speech
  recognition for indian languages,''
\newblock in {\em 2020 IEEE International Conference on Acoustics, Speech and
  Signal Processing}. IEEE, 2020, pp. 8279--8283.

\bibitem{zhou2022configurable}
Long Zhou, Jinyu Li, Eric Sun, and Shujie Liu,
\newblock ``A configurable multilingual model is all you need to recognize all
  languages,''
\newblock in {\em 2022 IEEE International Conference on Acoustics, Speech and
  Signal Processing}. IEEE, 2022, pp. 6422--6426.

\bibitem{zhu2020multilingual}
Yun Zhu, Parisa Haghani, Anshuman Tripathi, Bhuvana Ramabhadran, Brian Farris,
  Hainan Xu, Han Lu, Hasim Sak, Isabel Leal, Neeraj Gaur, et~al.,
\newblock ``Multilingual speech recognition with self-attention structured
  parameterization.,''
\newblock in {\em Proc. Interspeech}, 2020, pp. 4741--4745.

\bibitem{li2021prefix}
Xiang~Lisa Li and Percy Liang,
\newblock ``Prefix-tuning: Optimizing continuous prompts for generation,''
\newblock {\em arXiv preprint arXiv:2101.00190}, 2021.

\bibitem{graves2006connectionist}
Alex Graves, Santiago Fern{\'a}ndez, Faustino Gomez, and J{\"u}rgen
  Schmidhuber,
\newblock ``Connectionist temporal classification: labelling unsegmented
  sequence data with recurrent neural networks,''
\newblock in {\em Proceedings of the 23rd international conference on Machine
  learning}, 2006, pp. 369--376.

\bibitem{gales2014speech}
Mark~JF Gales, Kate~M Knill, Anton Ragni, and Shakti~P Rath,
\newblock ``Speech recognition and keyword spotting for low-resource languages:
  Babel project research at cued,''
\newblock in {\em Fourth International workshop on spoken language technologies
  for under-resourced languages}. ISCA, 2014, pp. 16--23.

\bibitem{conneau2020unsupervised}
Alexis Conneau, Alexei Baevski, Ronan Collobert, Abdelrahman Mohamed, and
  Michael Auli,
\newblock ``Unsupervised cross-lingual representation learning for speech
  recognition,''
\newblock {\em arXiv preprint arXiv:2006.13979}, 2020.

\bibitem{heafield2011kenlm}
Kenneth Heafield,
\newblock ``{K}en{LM}: Faster and smaller language model queries,''
\newblock in {\em Proceedings of the Sixth Workshop on Statistical Machine
  Translation}. 2011, pp. 187--197, Association for Computational Linguistics.

\end{thebibliography}

\end{document}